# Probabilistic graphs using coupled random variables

Kenric P. Nelson*[a], Madalina Barbu[b], Brian J. Scannell[c]
[a]Raytheon Corporation, 5 Apple Hill Rd., Tewksbury, MA, USA 01876-0345; [b]Raytheon Corporation, 125 President's Way, Woburn, MA 01801; Nanigans, 60 State St, 12th Floor Boston, MA 02109

## ABSTRACT

Neural network design has utilized flexible nonlinear processes which can mimic biological systems, but has suffered from a lack of traceability in the resulting network. Graphical probabilistic models ground network design in probabilistic reasoning, but the restrictions reduce the expressive capability of each node making network designs complex. The ability to model coupled random variables using the calculus of nonextensive statistical mechanics provides a neural node design incorporating nonlinear coupling between input states while maintaining the rigor of probabilistic reasoning. A generalization of Bayes rule using the coupled product enables a single node to model correlation between hundreds of random variables. A coupled Markov random field is designed for the inferencing and classification of UCI's MLR 'Multiple Features Data Set' such that thousands of linear correlation parameters can be replaced with a single coupling parameter with just a (3%, 4%) reduction in (classification, inference) performance.

## 1. INTRODUCTION

What if the nonlinear functions of neural networks, which provide much of their computational power, could be expressed within the constraints of a probabilistic graph? In this paper, we seek to show that the deformed algebra of nonextensive statistical mechanics offers an approach to this objective.

Artificial neural network (ANN) design has sought to increase the ability of computers to learn fundamental tasks in pattern recognition and control. While inspired by the functioning of biological neural networks, accurate models and comparable performance are still being sought. Beginning with the perceptron[1], which models a weighted sum of inputs and a bias followed by a nonlinear switch, the three fundamental components of an artificial neuron are conditioning of the input, fusion of a large collection of inputs, and a nonlinear function at the output. The single-layer perceptron is capable of linear discrimination. Multiple layers and backpropagation enabled the modeling of fundamental functions such as the XOR[2]. The introduction of nonlinear kernels, such as those used in support vector machines, greatly expanded the computational power and pattern recognition capabilities of ANNs.

Despite the computational power of neural network, trust in the application to real-world systems has continued to be limited by uncertainty about tracing the learned functions of complex nonlinear network to computational building blocks. The application of probabilistic graphs to the design of neural networks has sought to constrain the network design such that each node models a specific random variable, thereby adding control and traceability to the design. Furthermore, by restricting the nodes to well-defined random variables, modules can be systematically combined into higher-levels of reasoning. Deep belief networks[3], such as those introduced by G. Hinton, segment learning into layers which produce generative models for training of higher layers.

The challenge with probabilistic methods is that constraining each node to fusion using Bayes' rule, requires the design to either make assumptions about independence between random variables or significantly increase the complexity of network connectivity. The deformed algebra[4] of nonextensive statistical mechanics (NSM)[5], developed by

*kenric_p_nelson@raytheon.com; phone 1 603 508 9827;

Borges and others, extends probabilistic reasoning to a family of distributions associated with nonlinear processes. In particular a generalized product function can be used as a model of long-range dependence between random variables[6]. Factoring random variables, which for a Bayesian network requires identification of conditionally independence, can now be done by tuning a degree of nonlinear coupling between the variables. Quite general kernel functions at the input and sigmoid functions at the output can also be defined within this pseudo algebra; however, the focus of this paper will be on demonstrating the utility of fusion with these nonlinear functions.

In Section 2 the concept of a coupled random variable is introduced using an interpretation of NSM called nonlinear statistical coupling. This interpretation defines the generalized algebra in terms of the degree of nonlinear deformation. In Section 3 the Risk Profile is introduced, which uses the methods of nonlinear statistical coupling to measure the accuracy of inference as a function of risk. In Section 4 the design of a coupled Markov random field is demonstrated where the dependence structure is now global rather pair-wise. In Section 5 the results of a computational experiment using recognition of handwritten numerals are discussed.

## 2. DEFINITION OF A COUPLED RANDOM VARIABLE

The definition of a coupled random variable originates from the generalized product of nonlinear statistical coupling[7], which is an interpretation of nonextensive statistical mechanics. With this interpretation the statistical states of a statistical distribution are modeled as being coupled by nonlinear dynamics, rather than being mutually exclusive. Rather than measuring the probability of a state $p_i$, the coupled probability of a coupled state is measured by

$$P_i^{(\kappa)} = \frac{p_i \prod_{j=1, j \neq i}^{N} p_j^{\kappa}}{\sum_{i=1}^{N} p_i \prod_{j=1, j \neq i}^{N} p_j^{\kappa}} = \frac{p_i^{1-\kappa}}{\sum_{i=1}^{N} p_i^{1-\kappa}} \ . \tag{1}$$

The coupling $\kappa$ is a real-valued parameter and the non-coupled probability is recovered for $\kappa \to 0$. The middle expression makes explicit that the probability being measured is a coupling of the *i*th state with all the other states. The entropy measure is deformed using a coupled logarithm $\ln_{\kappa} x \equiv \frac{x^{\kappa} - 1}{\kappa}$ and averaged using the coupled probability

$$S_{\kappa} \equiv -\sum_{i=1}^{N} P_i^{(\kappa)} \ln_{\kappa} p_i \cdot , \tag{2}$$

Equation (2) is the normalized Tsallis entropy[8] expressed in terms of the degree of nonlinear coupling. Constraints on the coupled-mean and coupled-variance

$$\mu_{\kappa} \equiv \sum_{i=1}^{N} x_i p_i^{1-\kappa}; \quad \sigma_{\kappa}^2 \equiv \sum_{i=1}^{N} x_i^2 p_i^{1-\kappa} \tag{3}$$

lead to a maximum entropy distribution, referred to as a coupled Gaussian

$$p_i(x_i) = A_{\kappa} \exp_{\kappa} \left( \frac{-(x_i - \mu_{\kappa})^2}{(2+\kappa)\sigma_{\kappa}^2} \right) \ .$$
$$\exp_{\kappa}(x) \equiv e_{\kappa}^{x} = (1 + \kappa x)^{1/\kappa} \tag{4}$$

The coupled exponential is the inverse of the coupled logarithm.

Combinations of coupled exponentials and coupled logarithms lead to generalization of the product and sum functions

$$\mathrm{e}_\kappa^x \otimes_\kappa \mathrm{e}_\kappa^y \equiv \mathrm{e}_\kappa^{x+y} = \left(\left(\mathrm{e}_\kappa^x\right)^\kappa + \left(\mathrm{e}_\kappa^y\right)^\kappa - 1\right)^{1/\kappa}$$

$$\ln_\kappa x \oplus_\kappa \ln_\kappa y \equiv \ln_\kappa xy = \ln_\kappa x + \ln_\kappa y + \kappa\left(\ln_\kappa x\right)\left(\ln_\kappa y\right).$$

(5)

While the coupled product and coupled sum both satisfy associative and commutative properties, a distributive property cannot be defined, so these functions are part of a pseudo-algebra, which facilitates reasoning about the coupled states. The coupled sum makes explicit the non-additive property of the coupled entropy. The coupled division and coupled subtraction can also be defined. This discussion is limited to reviewing the properties of the coupled product, but full utilization of a probabilistic graph would include inferences that require division for independent variables and coupled-division for coupled variables. The coupled product defines a method for combining distributions which will be utilized to define a fusion of information referred to as a coupled random variable. For the purposes of this investigation defining a coupled random variable as one in which the distributions of the random variables are combined by the coupled product is sufficient to show an efficient model of correlation. Stronger definitions, using the definition of generalized independence[9] by Umarov, et. al. which uses the generalized product in a generalized Fourier domain and proved to form a generalized central limit theorem is desirable.

### 3. PROBABILITY ACCURACY AS A FUNCTION OF RISK

The methods of nonlinear statistical coupling provide a metric for measuring the accuracy of probability inferences, which is called a *Risk Profile*[10]. The Risk Profile measures the generalized mean of the true class probabilities. The adjustment in the mean is associated with risk due to the connection with the negative coupled logarithm, which modifies the cost of information. The generalized mean can be derived from the coupled entropy (1.2) by applying the coupled exponential, which transforms the entropy measure back into probability space

$$P_{avg} \equiv \exp_\kappa(-S_\kappa) = \exp_\kappa\left(\log_\kappa \prod_{i=1}^N {}_{\otimes_\kappa} p_i^{\otimes_\kappa^{P_i^{(\kappa)}}}\right) = \prod_{i=1}^N {}_{\otimes_\kappa} p_i^{\otimes_\kappa^{P_i^{(\kappa)}}}$$
$$= \prod_{i=1}^N {}_{\otimes_\kappa} \left(P_i^{(\kappa)} p_i^\kappa - \left(P_i^{(\kappa)} - 1\right)\right)_+^{1/\kappa} = \left(\sum_{i=1}^N P_i^{(\kappa)} p_i^\kappa\right)^{1/\kappa}.$$

(6)

The derivation uses the properties that sums of coupled logarithms are equal to the coupled logarithm of the coupled product of the arguments; that multiplying a coupled logarithm is equal to raising the argument to the coupled power, $x^{\otimes_\kappa^a} \equiv \left(ax^\kappa - (a-1)\right)^{1/\kappa}$; and that $\sum_{i=1}^N \left(P_i^{(\kappa)} - 1\right) = 1 - N$. For the Risk Profile, each test sample has equal weight $P_i^{(\kappa)} = 1/N$ and reports a probability for the true class $p_{i,true}$ resulting in the generalized mean of the reported probabilities for the true class

$$P_{accuracy} = \left(\frac{1}{N}\sum_{i=1}^N p_{i,true}^\kappa\right)^{1/\kappa}$$

(7)

For $\kappa \to 0$ the expression reduces to the geometric mean $P_{acc,\kappa\to 0} = \left(\prod_{i=1}^N p_{i,true}\right)^{1/N}$. Importantly, optimizing the accuracy of an inference, according to Shannon information theory, requires optimizing the geometric mean of the reported probabilities. Positive risk bias or nonlinear coupling, provides metrics which lower the cost of information and are similar to the finite costs on making a decision. Negative risk bias increases the cost of information, which is useful in evaluating the robustness of an inference algorithm. Only $\kappa \to 0$ is a proper score[11]; however, the biasing with risk provides useful insights and if necessary the full profile can be made proper, following the methods of the Tsallis proper score[12].

### 4. DESIGN OF A COUPLED MARKOV RANDOM FIELDS

Two prototypical probabilistic graphs are Bayesian networks and Markov networks[13]. Each is a model of a joint distribution showing the dependency relationships between the individual random variables. Bayesian networks are directed graphs which show the direction of causality. The children of a parent node represent the conditionally independent random variables dependent on the parent. The strength of the dependence is represented in conditional probability tables for each node. A Markov random field is an undirected graph with a weighted dependency between pairs of random variables.

The generalization utilizing nonlinear statistical coupling is relevant for both forms of probabilistic graphs. For a Bayesian network with child variable *A* and parent's *B* and *C*, Bayes rule for conditional independence

$$P(A|B,C) = P(B|A)P(C|A)P(A)/P(B,C) \qquad (8)$$

is replaced by a Coupled Bayes rule modeling a nonlinear dependence between the conditional probabilities

$$P(A_i | B \otimes_\kappa C) = \frac{\left[P^\kappa(B|A_i) + P^\kappa(C|A_i) - 1\right]^{1/\kappa} P(A_i)}{\sum_i \left[P^\kappa(B|A_i) + P^\kappa(C|A_i) - 1\right]^{1/\kappa} P(A_i)} \qquad (9)$$

The symbol $P(A_i | B \otimes_\kappa C)$ is used to represent that the posterior probability is based on the conditional probabilities given *B* and *C* are coupled. The function is written with respect to the *i*th class of *A* to show the normalizing sum of the coupled evidence. A complete design of this type of network requires consideration of how the conditional probability tables are modified given the coupled Bayes rule.

A coupled Markov random field provides a simpler illustration of how the use of coupled random variables provides an efficient model of dependence for complex systems. Taking as an example a multivariate normal distribution, the Markov random field shown in Figure 1 with dependencies shown as bold solid lines, is

$$P(A,B,C,D,E) = \frac{1}{Z} \exp\left(-\frac{1}{2} \sum_{i,j \in \mathcal{A}} x_i \begin{pmatrix} w_{AA} & w_{AB} & w_{AC} & 0 & 0 \\ w_{AB} & w_{BB} & 0 & 0 & 0 \\ w_{AC} & 0 & w_{CC} & w_{CD} & w_{CE} \\ 0 & 0 & w_{CD} & w_{DD} & 0 \\ 0 & 0 & w_{CE} & 0 & w_{EE} \end{pmatrix} x_j \right). \qquad (10)$$

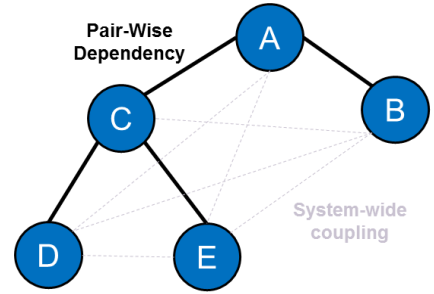

Figure 1. A Markov network defines pair-wise dependencies between random variables (bold). *Statistical coupling* defines a nonlinear dependency between a group of random variables (grey).

The matrix is the inverse of the covariance matrix and shows the weights of non-zero symmetric dependence. The means are set to zero and *Z* is the normalization of the distribution. The power and utility of the Markov random field is that additive exponent terms are also factorable into products of exponentials which facilitates inferencing and other probabilistic analysis. Unfortunately, as the size of a system grows inferencing becomes computationally intractable. While approximations such as Markov chain Monte Carlo are available, simpler models of the dependency are desirable. We next show that nonlinear statistical coupling can be used to approximate the dependency in a Markov random field with hundreds of random variables[14]. For the illustration in Figure 1 the coupled Markov random field forms a coupled multivariate Gaussian

$$P(A \otimes_\kappa B \otimes_\kappa C \otimes_\kappa D \otimes_\kappa E) = \frac{1}{Z} \exp_\kappa \left( \frac{-1}{2+\kappa} \sum_{i,j \in \mathcal{A}} x_i \begin{pmatrix} w_{AA} & 0 & 0 & 0 & 0 \\ 0 & w_{BB} & 0 & 0 & 0 \\ 0 & 0 & w_{CC} & 0 & 0 \\ 0 & 0 & 0 & w_{DD} & 0 \\ 0 & 0 & 0 & 0 & w_{EE} \end{pmatrix} x_j \right). \qquad (11)$$

In this model the pair-wise dependencies are set to zero and the additive arguments of the coupled exponential can be factored by the coupled product due to the properties of Equation (5). The coupling term can be chosen to approximate the joint distribution of Equation (10).

## 5. COUPLED MARKOV MODEL OF THE MULTIPLE FEATURES DATA SET

To illustrate the utility of a probabilistic graph using a coupled Markov random field the Machine Learning Repository's Multiple Features Data Set[15] is used. This dataset consists of 2000 handwritten numerals in which six feature sets have been extracted. Three models for fusion of the feature set will be compared; a multivariate normal distribution, single variate normal distributions fused using naïve Bayes, and the coupled Markov random field; i.e. single variate normal distributions fused using the coupled product. Table 1 shows the number of parameters, percent correct classification, and the probability accuracy for each feature set and each model. The multivariate model which requires a matrix of parameters grows as $n$-squared; whereas the coupled Markov model adds only one coupling parameter beyond the mean and variance for each feature.

For each feature set 1000 samples (100 per numeral) are used to calculate the mean and variance for each feature and a separate set of 1000 samples are used for testing. In the case of the multivariate normal the covariance parameters are also calculated. For the coupled fusion model, the accuracy measured by the geometric mean is maximized, examining coupling values between 0 and -2 in increments of -0.05. This is shown in Figure 3a.

The classification performance is similar for each fusion model. Assuming independence (naïve Bayes) or using the coupling method reduces the classification performance by 2-10%. The ability to maintain reasonable classification performance is why the assumption of independence is often acceptable. However, for resource management, fusion with other sources of information, or opportunities to defer decisions in order to collect additional information,[16] accurate probabilities are important for managing uncertainty. The accuracy of the probabilities should be evaluated using the geometric mean of the true class probabilities, given its connection to Shannon surprisal as explained in Section 3. This metric shows that the overconfidence of naïve Bayes reduces the accuracy of the probabilities. However, fusion using the coupled product restores the accuracy of the probabilities while maintaining a significantly less complex graphical model. Figure 2 shows graphical the ability of the coupled Markov random field to maintain the accuracy of the probabilities while significantly reducing the model complexity. The Principal Component Analysis, which are by design independent, is the only feature set with reasonable accuracy for the naïve Bayes model and no improvement using the coupled fusion. Accuracy similar to the full covariance matrix may be achievable for smaller than -0.05 coupling.

Table 1. Table of the number of parameters, percent correct classification, and the probability accuracy for each feature set and each fusion model. The percent correct classification is reduce by a few percentage points using naïve Bayes or coupling. The probability accuracy is measured using the geometric mean of the true class probabilities. The accuracy of the probabilities suffers for naïve Bayes when the features are correlated. The coupling model has comparable accuracy with the multivariate.

| Feature Type & Number | # of Parameters, Percent Correct Classification, Probability Accuracy | | |
|---|---|---|---|
| | **Naïve Bayes** | **Coupled** | **Multivariate** |
| **Fourier Coefficients 76** | 152   76%   21% | 153   76%   41% | 5,852   81%   50% |
| **Profile Correlations 216** | 432   89%   3% | 433   89%   73% | 46,872   98%   88% |
| **Principal Comp. Analysis 64** | 128   90%   56% | 129   90%   56% | 4,160   95%   80% |
| **Pixel Averages 240** | 480   91%   9% | 481   91%   69% | 57,840   94%   65% |
| **Zernike Moments 47** | 94   73%   17% | 95   73%   44% | 2,256   83%   60% |
| **Morphology 6** | 12   70%   25% | 13   70%   40% | 42   72%   48% |

Figure 3b shows the Risk Profile for the pixel averages feature set. The metric plots the generalized mean of the true class probabilities (7) as the nonlinear coupling or risk bias is adjusted. The naïve Bayesian model has overconfident probabilities but good classification performance, which is reflected in the high accuracy given positive risk bias with a steep reduction in accuracy as the risk bias is reduced. Use of the full covariance matrix shifts the Risk Profile curve to the left increasing the zero-bias accuracy of the probabilities. Nevertheless, the training of 46,872 parameters for this model runs the risk of being over fit to the data, which is reflected in the reduction in accuracy with negative risk bias. The coupled fusion is optimized for accuracy with no risk bias. Since only 433 parameters are used for this model there is less risk of over fitting reflected in the significant improvement in accuracy with negative risk bias. The gain in robustness is at the cost of less decisive probabilities. In principal the coupling could be tuned to optimize other risk bias points depending on the requirements of the inference algorithm.

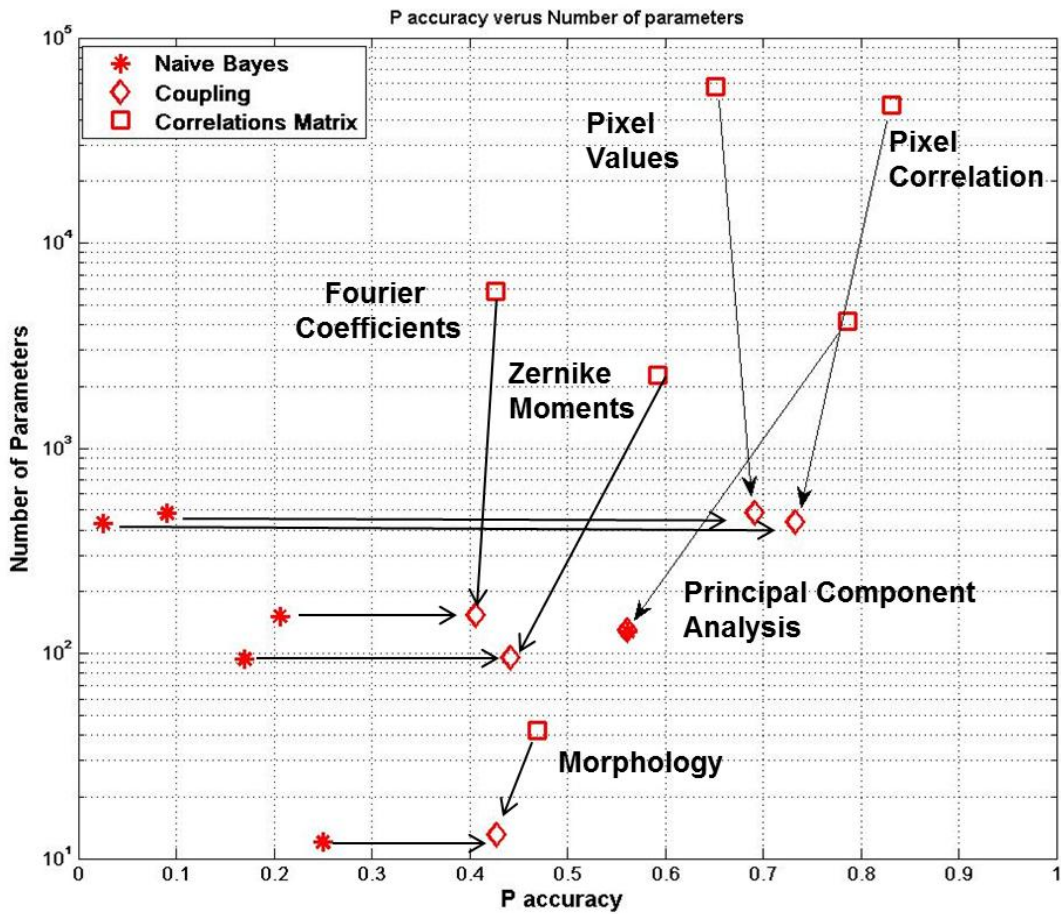

Figure 2. Plot of the number of parameters versus the probability accuracy for each feature fused using naïve Bayes (star), coupling (diamond), and full correlation matrix (square). The accuracy is measured using the geometric mean of the true class probabilities. Coupling uses only one more parameter than naïve Bayes, but has accuracy comparable to the full covariance matrix. The Principal Component Analysis is the only set with independent features, resulting in reasonable accuracy for naïve Bayes and no improvement with the coupling. Coupling smaller than the -0.05 increments tested may provide a modest.

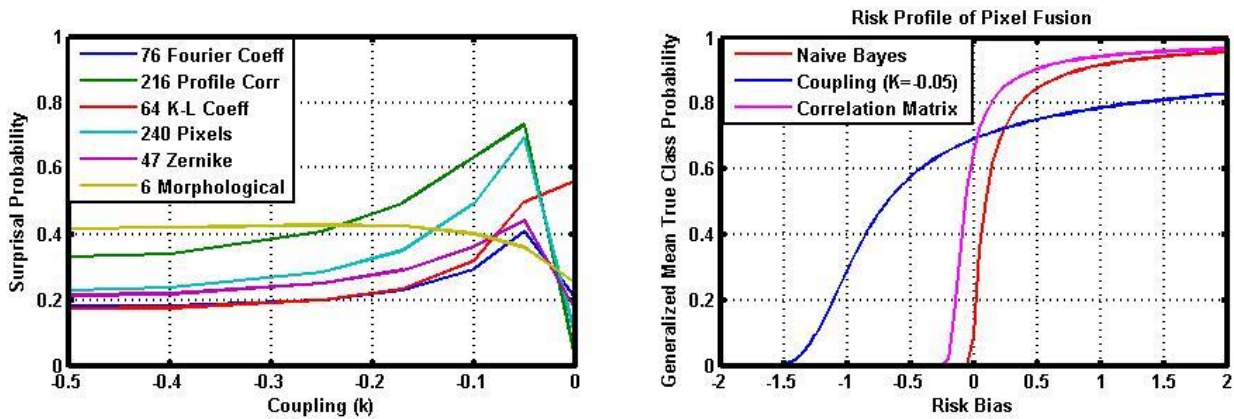

Figure 3. a) The coupling parameter for each feature set was selected by maximizing the geometric mean (or surprisal probability). b) The Risk Profile for the fusion of the pixel average features shows the characteristics of the three fusion methods. Naïve Bayes is decisive (positive risk bias), but inaccurate (zero risk bias). The correlation matrix is accurate but still not robust (negative risk bias). The coupling was optimized for accuracy and trades-off decisiveness for increased robustness.

## 6. CONCLUSION

This paper provides an approach to applying coupled random variables to the design of Bayesian and Markov networks. A numerical experiment using a coupled Markov random field to model the dependence between the features used for image processing of handwritten numerals is reviewed. The use of the coupled random variable adds one additional parameter to the naïve Bayesian model, but improves the probability accuracy to within a few percentage points of the full Markov correlation model. For the case of pixel value and pixel correlation features, which have 100s of feature values, there is 2 orders of magnitude reduction in the parameters between the full correlation and the coupling model.

Following information theoretic methods, the accuracy of probabilities is measured using the geometric mean of the true class probabilities, which is a translation into probability space of the measured cross-entropy between the model and the test samples. Further insight into the characteristics of the inference models is gained utilizing the cross-entropy of the normalized Tsallis entropy. Translated into probability space this metric is the generalized mean of the true class probabilities. Positive coupling or risk bias of this metric provides a measure of the decisive quality of the reported probabilities, while negative coupling or risk bias provides a measure of the robustness of the reported probabilities.

The three models tested show the following characteristics; the full Markov correlation model is decisive and accurate, but lacks robustness, which may be due to the large number of parameters necessary for the model; naïve Bayes is decisive, but inaccurate and non-robust, with the benefit of orders of magnitude fewer parameters; the coupled Markov model, which trains one coupling parameter to maximize the unbiased accuracy, does approximate the accuracy of the full Markov model, is more robust than either of the other models, but sacrifices decisiveness. Despite the lack of decisiveness in the coupled Markov model, the classification performance is equal to the naïve Bayes model. Furthermore, the improved accuracy provides system-level options to applying additional resources to those objects which report uncertainty.

Applying these methods to deep learning methodologies would be an important advance. The ability to factor the heavy-tail coupled-Gaussian using the coupled product, opens the possibility of developing networks which could generate heavy-tail samples for training of higher layers of the network.

## ACKNOWLEDGEMENTS

The authors benefited from discussion with Sabir Umarov and Fred Daum regarding modeling long range dependence with nonextensive statistical mechanics. The research was funded by a Raytheon IDEA project on Coupled Bayesian Networks.